\documentclass[letterpaper]{article} 
\usepackage{aaai2026}  
\usepackage{times}  
\usepackage{helvet}  
\usepackage{courier}  
\usepackage[hyphens]{url}  
\usepackage{graphicx} 
\usepackage{natbib}  
\usepackage{caption} 
\usepackage{algorithm}
\usepackage{algorithmic}

\usepackage{newfloat}
\usepackage{listings}

\usepackage{amsmath}
\usepackage{amssymb}
\usepackage{mathtools}
\usepackage{amsthm}
\usepackage{pifont}
\usepackage{booktabs, multirow}

\usepackage{newfloat}
\usepackage{listings}
\DeclareCaptionStyle{ruled}{labelfont=normalfont,labelsep=colon,strut=off} 
\floatstyle{ruled}
\newfloat{listing}{tb}{lst}{}
\floatname{listing}{Listing}
\title{MMPG: MoE-based Adaptive Multi-Perspective Graph Fusion for \\ Protein Representation Learning}

\author {
    Yusong Wang\textsuperscript{\rm 1}\equalcontrib,
    Jialun Shen\textsuperscript{\rm 2}\equalcontrib,
    Zhihao Wu\textsuperscript{\rm 3},
    Yicheng Xu\textsuperscript{\rm 2},
    Shiyin Tan\textsuperscript{\rm 2},
    Mingkun Xu\textsuperscript{\rm 1}\thanks{Corresponding author: xumingkun@gdiist.cn},\\
    Changshuo Wang\textsuperscript{\rm 4},
    Zixing Song\textsuperscript{\rm 5},
    Prayag Tiwari\textsuperscript{\rm 6}
}
\affiliations {
    \textsuperscript{\rm 1}Guangdong Institute of Intelligence Science and Technology, Zhuhai, China\\
    \textsuperscript{\rm 2}Institute of Science Tokyo, Tokyo, Japan\\
    \textsuperscript{\rm 3}Zhejiang University, Hangzhou, China\\
    \textsuperscript{\rm 4}University College London, London, U.K.\\
    \textsuperscript{\rm 5}University of Cambridge, Cambridge, U.K.\\
    \textsuperscript{\rm 6}Halmstad University, Halmstad, Sweden\\
}

\usepackage{bibentry}

\begin{document}

\maketitle

\begin{abstract}
Graph Neural Networks (GNNs) have been widely adopted for Protein Representation Learning (PRL), as residue interaction networks can be naturally represented as graphs.
Current GNN-based PRL methods typically rely on single-perspective graph construction strategies, which capture partial properties of residue interactions, resulting in incomplete protein representations.
To address this limitation, we propose MMPG, a framework that constructs protein graphs from multiple perspectives and adaptively fuses them via Mixture of Experts (MoE) for PRL.
MMPG constructs graphs from physical, chemical, and geometric perspectives to characterize different properties of residue interactions.
To capture both perspective-specific features and their synergies, we develop an MoE module, which dynamically routes perspectives to specialized experts, where experts learn intrinsic features and cross-perspective interactions.
We quantitatively verify that MoE automatically specializes experts in modeling distinct levels of interaction—from individual representations, to pairwise inter-perspective synergies, and ultimately to a global consensus across all perspectives.
Through integrating this multi-level information, MMPG produces superior protein representations and achieves advanced performance on four different downstream protein tasks.
\end{abstract}


\section{Introduction}

Protein Representation Learning (PRL) has emerged as a fundamental methodology in bioinformatics, aiming to encode the structural, biochemical, and functional properties of proteins into informative representations. 
These learned representations are crucial across a wide range of downstream applications, including drug discovery, functional annotation, and protein design \citep{Hoang_Sbodio_Martinez}.
Since protein properties arise from complex interactions among amino acids (residues), it is essential for PRL methods to capture these patterns. 
A natural way to model such residue-level interactions is through graphs, where residues are represented as nodes and their relationships (e.g., spatial proximity or chemical bonds) as edges. 
Therefore, Graph Neural Networks (GNNs) have been widely adopted for PRL, as they enable message passing and aggregation across residues to learn expressive protein representations \citep{4700287,zhang2023protein,wang2025enhancing}.

\begin{figure}
\includegraphics[width=0.465\textwidth]{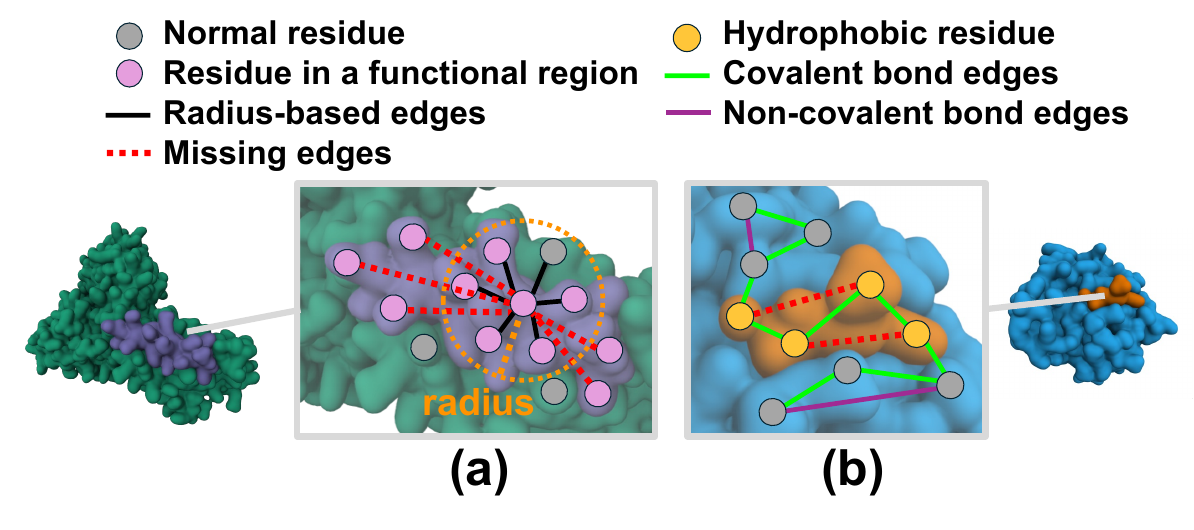}
\centering
\caption{
Limitations of protein graph construction from a single perspective.
(a) A radius-based graph misses long-range connections.
(b) A chemical-bond-based graph fails to capture the association of adjacent hydrophobic residues.
}
\label{fig:weakness_example}
\end{figure}

GNN-based PRL methods have achieved remarkable success in recent years \cite{wang2023learning, zhang2023protein, wang2025enhancing}.
They vary in their protein graph construction strategies with specialized GNN architectures to capture protein properties effectively.
For instance, ProNet \citep{wang2023learning} leverages radius-based graphs with hierarchical encoders for protein structural modeling, excelling in predicting enzyme reaction type.
GearNet \citep{zhang2023protein} stacks sequential, radius, and K-Nearest Neighbors (KNN) edges in a graph to capture geometric proximity relationships of residues at different scales, and then employs a relational Graph Convolutional Network (GCN) to learn protein graph representation, achieving strong performance across multiple protein tasks.
However, these methods are limited to single-perspective graph construction, capturing only one semantic aspect of residue interactions.
As illustrated in Figure~\ref{fig:weakness_example} (a), a radius-based graph may miss functionally coupled distant residues, and a chemical-bond-based graph might overlook adjacent hydrophobic residues driven by the hydrophobic effect (Figure~\ref{fig:weakness_example}(b)) \citep{doi:10.1126/science.653353}.
This loss of information limits the expressive power of learned protein representations, constraining their performance at downstream protein tasks.

These limitations motivate us to explore a multi-perspective-based representation learning method. 
We begin by constructing a set of graphs, each encoding a distinct semantic perspective of residue interactions: 
1) a physical-energetic perspective capturing interaction stability via knowledge-based potentials; 
2) a chemical-functional perspective encoding residue similarities based on biochemical properties; 
and 3) a geometric-structural perspective modeling local spatial relationships.
Critically, these perspectives are not isolated but exhibit complex interdependencies. 
Each perspective offers unique information and their synergistic interactions form the basis for a comprehensive understanding of protein properties.
For example, a geometric graph identifies approximate neighbor residues, and a physical-energetic graph helps validate the stability of their interactions. 
To learn both information, we develop a Mixture-of-Experts (MoE) module.
Through automatic routing of perspectives to experts, each expert learns different knowledge ranging from intra-perspective features to inter-perspective synergies.
Our analysis of expert selection frequencies verifies that MoE specializes its experts to capture interaction levels, ranging from individual representations to pairwise inter-perspective synergies, and ultimately a global consensus across all perspectives.
Through this \textbf{\underline{M}}oE‑based adaptive \textbf{\underline{M}}ulti‑\textbf{\underline{P}}erspective \textbf{\underline{G}}raph fusion framework, MMPG produces expressive protein representations.

Main contributions are summarized as:
\begin{itemize}\setlength{\itemsep}{0pt}
\item We construct protein graphs from three semantic domains of physical, chemical, and geometric perspectives, providing comprehensive coverage of residue interactions beyond single-perspective limitations.
\item  We develop an MoE module that discovers and leverages multi-level interactions among perspectives to achieve effective multi-perspective information integration.
\item  We provide quantitative evidence that MoE module can specialize its experts to capture cross-perspective interactions at multiple levels. This multi-perspective integrating mechanism enables MMPG to achieve advanced performance on four different downstream protein tasks, demonstrating its effectiveness on PRL.

\end{itemize}

\section{Related Work}
\subsection{Protein Representation Learning} 
PRL is commonly categorized into sequence-based and structure-based approaches.
Sequence-based methods use architectures like word embeddings to capture individual residue features \citep{Yang2018LearnedPE}, 1D-CNNs to extract local sequence motifs \citep{Kulmanov2019DeepGOPlusIP}, and Transformers to model long-range dependencies between residues \citep{doi:10.1126/science.ade2574}.
Structure-based approaches, mainly based on GNNs, explicitly capture structural information, such as spatial relationships between residues, for a more comprehensive understanding of protein structures \citep{zhang2023protein,wang2023learning,jamasb2024evaluating,wang2025enhancing}.
The performance of GNN-based methods depends heavily on the initial protein graph construction, which remains a key bottleneck.

\subsection{Protein Graph Construction} 
Protein graph construction abstracts a protein's 3D conformation into a graph.
In this work, we focus on residue-level graphs where a node corresponds to a residue.
The residue interaction network is reflected in the edge connectivity patterns \citep{Fasoulis2021GraphRL}.
Constructing protein graphs is non-trivial, as multiple chemical, physical, and geometric properties of proteins should be considered. 
Based on these properties, various methods are developed.
Geometric-based methods, such as radius, KNN, or Delaunay triangulation, are used to capture spatial relationships \citep{Quan2024ClusteringFP, wang2025enhancing, jamasb2024evaluating}. 
Chemical-based methods define edges based on chemical relationships of residues, such as chemical bonds \citep{Baldassarre2020GraphQAPM}.
Most PRL methods adopt only one perspective, thereby partially encoding residue interactions.
Few attempts try multi-perspective construction, merely adding another perspective on top of a primary one \citep{Baldassarre2020GraphQAPM, CHIANG2022100975}.
They use coarse fusion strategies, such as edge stacking or perspective feature concatenation \citep{Baldassarre2020GraphQAPM}, where indiscriminate merging of cross-perspective information introduces noise and dilutes key structural signals.
Instead, MMPG constructs protein graphs from physical, chemical, and geometric perspectives, using an MoE model to adaptively fuse their information, yielding richer protein representations.

\subsection{Mixture of Experts} 
Recent advances highlight MoE's effectiveness in capturing both shared and specialized data patterns. In multi-task learning, MoE is introduced to balance common and task-dependent representations \citep{10.1145/3219819.3220007,Chen2023ModSquadDM}. 
In multi-modal learning, \citet{10777289} and \citet{liang2025mixtureoftransformers} employ MoE to disentangle modality-shared/specific patterns, enabling more effective cross-modal interaction modeling.
Motivated by its success, we explore MoE in PRL as a principled module to capture both shared and specific information from multi-perspective protein graphs.

\begin{figure*}
\includegraphics[width=0.98\textwidth]{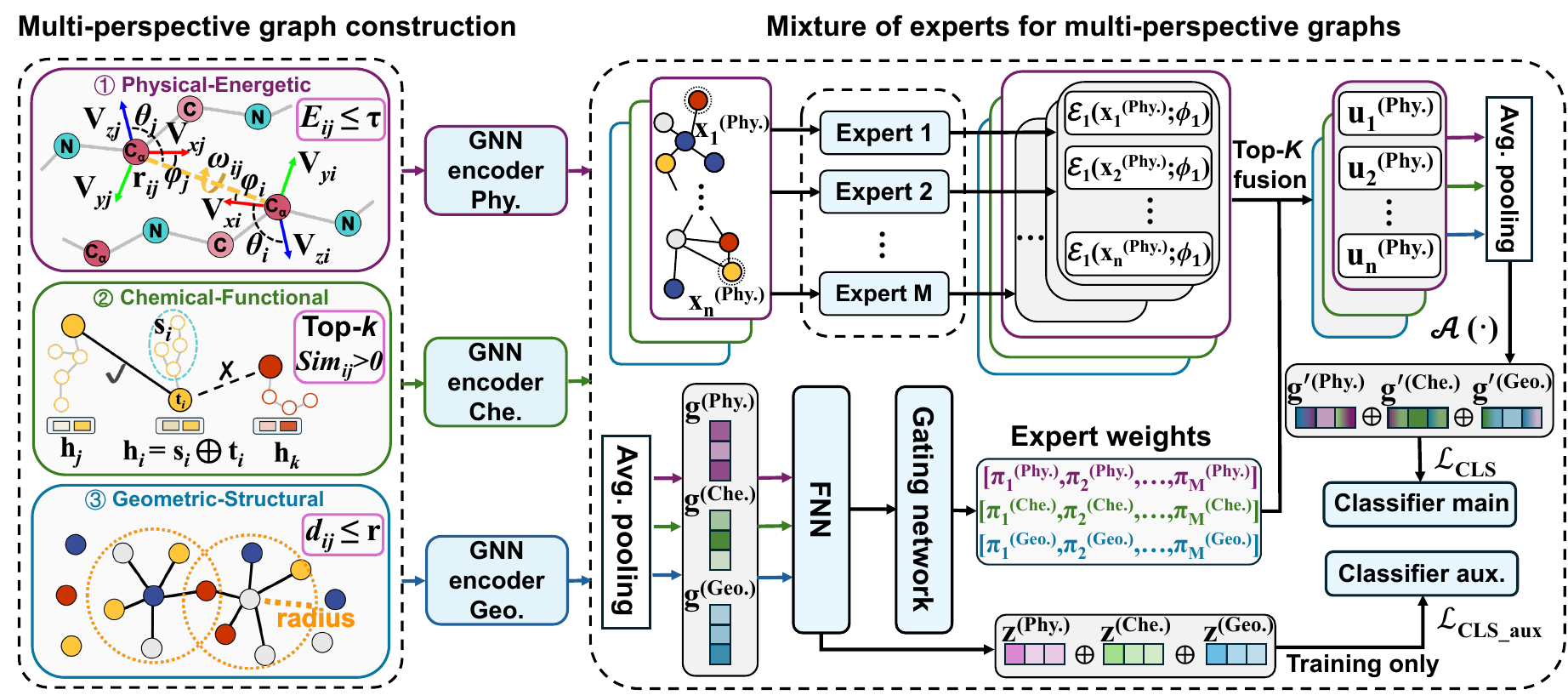}
\centering
\caption{
Overview of the proposed MMPG framework, which consists of two stages: (1) Multi-Perspective Graph Construction. Three graphs are constructed to model physical, chemical, and geometric properties of residue interaction, and (2) MoE learning scheme. Perspectives are routed to specialized experts, enabling dynamic representation learning across different perspectives.
}
\label{fig:structure}
\end{figure*}

\section{Preliminary}

\textbf{Notations \& problem definition.}
Let $G=(\mathcal{V}, \mathcal{E}, \mathcal{P})$ be a protein graph, where $\mathcal{V} = \{v_i\}_{i=1}^{n}$ denotes the set of nodes (residues, each represented by its central carbon atom $\mathrm{C}_{\alpha}$), and $\mathcal{E} = \{e_{ij}\}$ represents edge set. 
$\mathcal{P} = \{\mathbf{p}_i\}_{i=1,\dots,n}$ denotes coordinate set, where each $\mathbf{p}_i \in \mathbb{R}^{3 \times 1}$ represents the coordinate of node $i$.
Node features are a concatenation of learned embeddings for both residue type and side-chain conformation.
Edge features are built from a rotationally invariant relative spatial encoding scheme \citep{fan2023continuousdiscrete}, combining sequence information with normalized relative position vectors, projections in the local coordinate system, and pairwise distances.
Objective of protein prediction is to learn a mapping $f: G \to \mathcal{Y}$, where $\mathcal{Y}$ is the set of labels. 

\noindent\textbf{Graph encoder.}
All graph encoders in our framework use an edge‑aware GCN \citep{wang2025enhancing}.
The hidden feature of node~$i$ is updated from layer~$l$ to layer~$l+1$ via:

{\fontsize{8}{14}
\begin{equation}
\begin{aligned}
\mathbf{h}_{i}^{(l+1)}
=\sigma\!\left(
  \sum_{j\in\mathcal{N}(i)}
  \frac{1}{\sqrt{\lvert\mathcal{N}(i)\rvert\,\lvert\mathcal{N}(j)\rvert}}\;
  \bigl(\mathbf{h}_{j}^{(l)}\mathbf{W}_{n}\bigr)
  \odot
  \!\bigl(\mathbf{e}_{ij}\mathbf{W}_{e}\bigr)
\right),
\label{eq:edge_gcn}
\end{aligned}
\end{equation}}

\noindent where $\mathbf{h}_{i}^{(l)}$ is the feature vector of node $i$ at layer $l$,
$\mathcal{N}(i)$ denotes the neighbors of $i$ 
,
$\mathbf{e}_{ij}$ is the edge feature between nodes $i$ and $j$,
$\mathbf{W}_{n}$ and $\mathbf{W}_{e}$ are learnable weights,	
$\odot$ denotes element‑wise product,
and $\sigma(\cdot)$ is the activation function.

\section{Methodology}
In this section, we first present our multi-perspective protein graph construction strategies.
We then introduce an MoE framework for integrating these multi-perspective protein graphs. 
The overall architecture is illustrated in Figure~\ref{fig:structure}.

\subsection{Multi-Perspective Protein Graph Construction}
\label{construction}

\noindent\textbf{Physical-Energetic Perspective.}
To endow our protein graphs with rich physical–energetic semantics, we adopt KORP~\citep{lopez2019korp}, a knowledge-based pairwise potential which quantitatively captures effective inter-residue interaction energies from experimentally solved structures, thus directly reflecting conformational preferences and stability determinants of proteins~\citep{sippl1995knowledge, miyazawa1996residue}.

KORP is grounded on a coarse-grained residue representation and a 6-D joint probability distribution that considers the relative orientation and position of residue pairs. 
As illustrated in the left top of Figure~\ref{fig:structure}, each residue pair $(i,j)$ is described by:
a distance parameter $\mathbf{r}_{ij}$;
four angular parameters $\theta_i, \phi_i, \theta_j, \phi_j$, which are the polar coordinates of $\mathbf{r}_{ij}$ in the local 3D reference frame of each amino acid, defined as:
\begin{equation}
\begin{aligned}
    \mathbf{V}_z &= (\mathbf{r}_{CC_\alpha}+\mathbf{r}_{NC_\alpha}) / |\mathbf{r}_{CC_\alpha}+\mathbf{r}_{NC_\alpha}|, \\
    \mathbf{V}_y &= (\mathbf{V}_z\times\mathbf{r}_{NC_\alpha}) / |\mathbf{V}_z\times\mathbf{r}_{NC_\alpha}|, \\
    \mathbf{V}_x &= (\mathbf{V}_y\times\mathbf{V}_z),
\end{aligned}
\end{equation}
where $\mathbf{r}_{CC_\alpha}=\mathbf{r}_C-\mathbf{r}_{C_\alpha}$ and $\mathbf{r}_{NC_\alpha}=\mathbf{r}_N-\mathbf{r}_{C_\alpha}$ are vectors from C$_\alpha$ to carbonyl carbon (C) and nitrogen (N) atoms of the same residue, respectively; 
torsional angle $\omega_{ij}$ describes relative rotation of vectors $\mathbf{V}_{zi}$ and $\mathbf{V}_{zj}$ along $\mathbf{r}_{ij}$ axis.
Potential for residue pair $(i,j)$ with types $(a,b)$ is given by inverse‐Boltzmann equation \citep{choulli1996inverse}: 
\begin{equation}
    E_{ij} = -RT\ln\frac{\text{P}^{obs}_{ab}(r_{ij}, \theta_i, \varphi_i, \theta_j, \varphi_j, \omega_{ij})+z}{\text{P}^{ref}(r_{ij}, \theta_i, \varphi_i, \theta_j, \varphi_j, \omega_{ij})+z},
    \label{eq:korp}
\end{equation}
where $R$ is the molar gas constant, 
$T$ is the temperature, 
$\text{P}^{obs}_{ab}$ is the joint probability at the given relative distance and orientation of observing two amino acids $i$ and $j$ of type $a$ and $b$, 
$\text{P}^{ref}$ is the reference probability using the classical reference state~\citep{samudrala1998all}, 
and $z$ is a smoothing constant to stabilize low-count statistics.
We connect nodes $i$ and $j$ if $E_{ij} \leq \tau$, where $\tau$ is a predefined threshold.
This yields a protein graph whose edges encode stable structural motifs and protein backbone rigidity, capturing critical long-range and non-covalent interactions (e.g., hydrophobic effects), thereby providing physical insights for PRL.

\noindent\textbf{Chemical-Functional Perspective.}
This perspective aims to model residue relationships based on chemical-functional similarities. 
Thus, we construct residue embeddings based on amino acid type and side chain conformation.
The amino acid type determines the residue's intrinsic biochemical attributes (e.g., polarity, charge) \citep{KYTE1982105}, and the side-chain conformation dictates the spatial presentation of chemical properties \citep{janin1978conformation}. 
They together define a residue's chemical semantics, determining its biological function, such as being part of a catalytic center.
The similarity among these embeddings identifies residues with analogous roles rather than those merely interacting.
We implement these two embeddings as follows:

\textit{Amino acid type embedding.} Each residue is characterized by its underlying amino acid type, which is mapped to a unique index and embedded into a learnable vector:
\begin{equation}
\mathbf{t}_i = \mathrm{Embedding}(\mathrm{type}_i).
\end{equation}
These embeddings learn to reflect the intrinsic chemical properties in a data-driven manner.
\textit{Side chain conformation embedding} is based on a residue's first four torsion angles $(\chi^1, \chi^2, \chi^3, \chi^4)$, which uniquely determine the side chain conformation given a fixed protein backbone~\citep{wang2023learning}. 
Each torsion angle uses a sine–cosine encoding:
\begin{equation}
\mathbf{s}_i = \text{concat} \left( [\sin \chi_i^n, \cos \chi_i^n] \right)_{n=1}^4,
\end{equation}
where the raw embedding $\mathbf{s}_i$ is subsequently projected by a linear layer to the same dimension as $\mathbf{t}_i$.
To encode their combined effect, $\mathbf{t}_i$ and $\mathbf{s}_i$ are concatenated and passed through a feed-forward neural network (FNN), yielding the chemical-functional embedding:
\begin{equation}
    \mathbf{h}_i = \mathrm{FNN}(\mathbf{t}_i \oplus \mathbf{s}_i).
\end{equation}
The similarity between residues $i$ and $j$ is then quantified by the cosine similarity of their state embeddings:
\begin{equation}
Sim_{ij} = \frac{\mathbf{h}_i^T \mathbf{h}_j}{\|\mathbf{h}_i\|_2 \|\mathbf{h}_j\|_2}.
\end{equation}
We construct edges using a hybrid thresholded top-$k$ strategy:
For each residue $i$, we first identify its $k$ most similar neighbors, and then form edges only with those neighbors $j$ with a positive similarity score ($\mathrm{Sim}_{ij} > 0$), ensuring both graph sparsity and semantically meaningful connections.

\begin{table*}[h]
\centering
\setlength{\tabcolsep}{4.16 mm}{
\begin{tabular}{l|c|ccc|ccc|c}
    \toprule
    \multirow{2}{*}{\textbf{Method}} &
    \multirow{2}{*}{\textbf{EC}} &
    \multicolumn{3}{c|}{\textbf{GO}} &
    \multicolumn{3}{c|}{\textbf{FOLD}} &
    \multirow{2}{*}{\textbf{Reaction}} \\
    & & \textbf{BP} & \textbf{MF} & \textbf{CC} & \textbf{Fold} & \textbf{Super.} & \textbf{Fam.} \\
    \midrule
    \multicolumn{9}{c}{\textit{Protein Representation Learning Methods}} \\
    \midrule
    3DCNN          & 0.077 & 0.240 & 0.147 & 0.305 & 31.6 & 45.4 & 92.5 & 72.2 \\
    GraphQA        & 0.509 & 0.308 & 0.329 & 0.413 & 23.7 & 32.5 & 84.4 & 60.8 \\
    ProtBERT-BFD        & 0.838 & 0.279 & 0.456 & 0.408 & 26.6 & 55.8 & 97.6 & 72.2 \\
    GVP            & 0.489 & 0.326 & 0.426 & 0.420 & 16.0 & 22.5 & 82.8 & 65.5 \\
    LM-GVP        & 0.664 & 0.417 &  0.545 & 0.527 & 48.3 & 70.3 & 99.5 & 85.3 \\
    DeepFRI            & 0.631 & 0.399 &  0.465 & 0.460 & 15.3 & 20.6 & 73.2 & 63.3 \\
    IEConv         &   -   & 0.421 & 0.624 & 0.431 & 47.6 & 70.2 & 99.2 & 87.2 \\
    ProNet         &   -   &   -   &   -   &   -   & 52.7 & 70.3 & 99.3 & 86.4 \\
    GearNet        & 0.810 & 0.400 & 0.581 & 0.430 & 48.3 & 70.3 & 99.5 & 85.3 \\
    ESM-2         & 0.861 & 0.460 & 0.662 & 0.427 & 38.5 & \textbf{81.5} & 99.2 & - \\
    CDConv         & 0.820 & 0.453 & 0.654 & 0.479 & 56.7 & 77.7 & \textbf{99.6} & 88.5 \\
    EPGGCL         & 0.885 & 0.454 & 0.659 & 0.477 & 59.8 & 80.8 & 99.5 & \textbf{89.0} \\
    \midrule
    \multicolumn{9}{c}{\textit{Protein Graph Construction Strategies}} \\
    \midrule
    Chemical bond       &   0.620    &  0.301  &  0.552 &  0.411  &   13.6    &  14.3    &  68.0    &  46.6   \\
    KNN  &    0.821   & 0.434 &   0.610   & 0.415 & 55.4 & 75.6 & 98.9 & 85.2 \\
    Radius              & 0.840 & 0.419 & 0.631 & 0.419 & 55.3 & 75.7 & 99.3 & 84.1 \\
    Delaunay triangulation &  0.774  &    0.332   &   0.567   &   0.424    &   24.9   &   47.0   &     94.2    &    81.6  \\
    \midrule
    MMPG             & \textbf{0.893} & \textbf{0.463} &   \textbf{0.663}    & \textbf{0.489} & \textbf{60.9} & 79.5 & \textbf{99.6} & \textbf{89.0} \\
    \bottomrule
\end{tabular}
}
\caption{
Comparison of MMPG with two different types of baselines across multiple protein-related tasks. ``-'' indicates results not reported in the original paper.
\textbf{Bold} shows the best performance.
}
\label{tab:comparison}
\end{table*}

\noindent\textbf{Geometric-Structural Perspective.}
We construct the graph based on geometric proximity to represent the local structural context of the protein.
Specifically, an edge is added between residues $i$ and $j$ if $d_{ij}\leq r$, where $d_{ij}$ is the Euclidean distance between their C$_\alpha$ atoms and $r$ is a predefined radius.
These geometric edges complement the sequential connectivity by capturing residues that may be distant in sequence but are brought into close proximity in the folded structure. 
This construction is critical for encoding the three-dimensional spatial relationships that are essential for protein function \citep{doi:10.1073/pnas.1102727108}.

After constructing graphs from three distinct perspectives, we apply a separate graph encoder to each, yielding node embeddings that capture perspective-specific features.

\subsection{Mixture of Experts for Multi-Perspective Graphs}
\label{moe}

To effectively integrate multi-perspective features, we leverage an MoE module to process three protein graphs. 
This design enables learning both perspective-specific information and cross-perspective interaction semantics through different experts, yielding comprehensive protein graph representations.
MoE consists of two components: 1) a shared pool of experts, where each expert implements a graph encoder to process graphs from any perspective, but through training naturally develops specialization for certain intra- or inter-perspective patterns; and 2) a gating network that produces weights for all experts based on the input perspective graph and routes the graph to the most suitable experts accordingly.

Given the graph-level embedding of a perspective $\mathbf{g}^{(p)} = \frac{1}{n} \sum_{i=1}^{n} \mathbf{h}_i^{(p)}$ (where $p$ denotes perspective), the gating weight of the $m$-th expert for perspective $p$ is:
\begin{equation}
\pi_m^{(p)}(\mathbf{g}; \boldsymbol{\theta}) = \frac{\exp(\mathcal{G}(\mathbf{g}^{(p)}; \boldsymbol{\theta})[m])}{\sum_{i=1}^{M} \exp(\mathcal{G}(\mathbf{g}^{(p)}; \boldsymbol{\theta})[i])}, \quad \forall m \in M,
\end{equation}
where $\mathcal{G}(\cdot)$ is the gating function, a one-layer FNN
parameterized by $\boldsymbol{\theta}$ that outputs a weight vector of dimension $M$; 
$[m]$ denotes its $m$-th element.
Following \citet{shazeer2017}, we adopt a top-$K$ strategy, selecting the $K$ experts with the highest gating weights for each input.
The node representation for perspective $p$ is a weighted sum of $K$ experts:
\begin{equation}
\mathbf{u}_i^{(p)} = \sum_{k=1}^{K} \pi_{k}^{(p)}\, \mathcal{E}_k(\mathbf{x}_i^{(p)}; \boldsymbol{\phi}_k),
\end{equation}
where 
$\mathcal{E}_k(\cdot)$ denotes the $k$-th expert parameterized by $\boldsymbol{\phi}_k$, $\mathbf{x}_i^{(p)}$ is the input feature of node $i$ in perspective $p$.
This weighted fusion allows each node to integrate information from both its perspective-specific experts and shared experts across perspectives, thereby capturing both unique and shared semantic features.
Furthermore, a load-balancing regularizer is imposed to encourage balanced expert selection~\cite{shazeer2017}, which is excluded from the task loss.
The updated graph-level embedding for each perspective is obtained by a mean pooling readout over all nodes:
\begin{equation}
\mathbf{g'}^{(p)} = 
\frac{1}{n} \sum_{i=1}^{n}\mathbf{u}_i^{(p)}.
\end{equation}

Finally, we use concatenation as aggregation function $\mathcal{A}(\cdot)$ to obtain protein representation $\mathbf{g}_{\text{fused}}$ from $P$ views:
\begin{equation}
\mathbf{g}_{\text{fused}} = \mathcal{A}\left(\left\{\mathbf{g'}^{(p)}\right\}_{p=1}^{P}\right).
\end{equation}

\subsection{Optimization}
MMPG is trained using a joint task loss that integrates both global and auxiliary supervision~\footnote{Following \citep{fan2023continuousdiscrete}, we use NLL loss for single-label protein classification tasks (RC and FOLD) and BCE loss for multi-label protein classification tasks (GO and EC).}:
\begin{equation}\label{Totalloss}
    \mathcal{L}_\mathrm{TASK} = \mathcal{L}_\mathrm{CLS} + \lambda \mathcal{L}_{\text{CLS\_aux}},
\end{equation}
$\mathcal{L}_\mathrm{CLS}$ provides global supervision to  $\mathbf{g}_{\text{fused}}$.
$\mathcal{L}_{\text{CLS\_aux}}$ provides auxiliary supervision to the concatenated FNN outputs $\text{concat}(\mathbf{z}^{(p)})_{p=1}^3$, which encourages the upstream GNN encoders to produce task-discriminative features that are then fed to the gating network for stable expert routing.
Coefficient $\lambda$ balances the contributions of the two losses.

\begin{table*}[h]
\centering
\setlength{\tabcolsep}{4.05 mm}{
\begin{tabular}{l|c|ccc|ccc|cc}
    \toprule
    {\multirow{2}{*}{\textbf{Method}}}
     & \multirow{2}{*}{\textbf{EC}} & \multicolumn{3}{c|}{\textbf{GO}} & \multicolumn{3}{c|}{\textbf{FOLD}} & \multirow{2}{*}{\textbf{Reaction}} \\ 
      &  & \textbf{BP} & \textbf{MF} & \textbf{CC} & \textbf{Fold} & \textbf{Super.} & \textbf{Fam.} \\ 
    \midrule
    w/o Physical-energetic & 0.874 & 0.448 & 0.652 & 0.466 & 59.0 & 77.4 & 99.5 & 84.9 \\
    w/o Chemical-functional & 0.851 & 0.440 & 0.635 & 0.454 & 57.0 & 76.7 & 99.3 & 86.5 \\
    w/o Geometric-structural & 0.821 & 0.422 & 0.623 & 0.436 & 52.0 & 72.1 & 99.0 & 84.3 \\
    \midrule
    w/o MoE (edge stack) & 0.794 & 0.372  & 0.582 & 0.446 & 54.5 & 75.7 & 99.3 & 85.6 \\
    w/o MoE (concatenation) & 0.865 & 0.426 & 0.600 & 0.452 & 56.9 & 76.2 & 99.3 & 84.9\\
    w/o $\mathcal{L}_{\text{CLS\_aux}}$ & 0.881 & 0.456 & 0.648 & 0.474 & 58.8 & 77.4 & 99.5 & 87.8\\
    \midrule
    Complete & \textbf{0.893} & \textbf{0.463} &   \textbf{0.663}    & \textbf{0.489} & \textbf{60.9} & \textbf{79.5} & \textbf{99.6} & \textbf{89.0} \\
    \bottomrule
\end{tabular}}
\caption{Ablation study evaluates the effectiveness of each component by removing the three semantic perspectives and the MoE fusion module one by one. The general drop in performance across all tasks demonstrates the necessity of each component.
}
\label{tab:ablation}
\end{table*}

\begin{figure*}[h]
\includegraphics[width=1\textwidth]{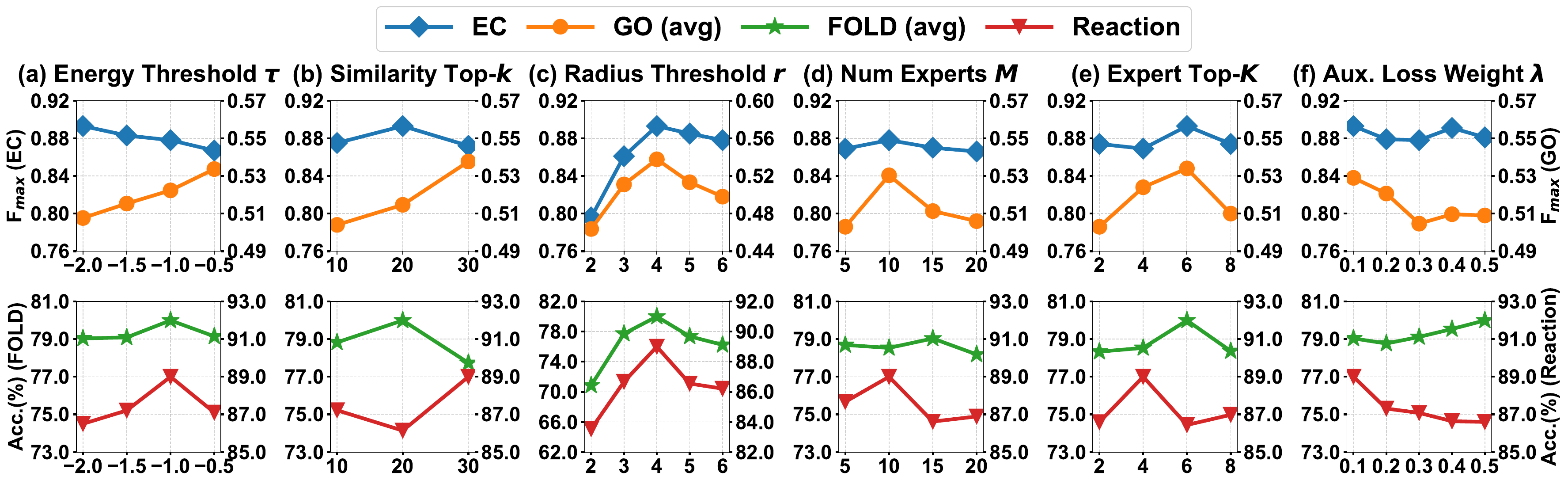}
\centering
\caption{
The plots show MMPG's performance as key hyperparameters for the graph construction and MoE module are varied.
}
\label{fig:param}
\end{figure*}

\section{Experiments}

\subsection{Experiment Setting}

\noindent\textbf{Tasks.} We evaluate on four representative protein tasks: (1) Protein Fold Classification (FOLD) \citep{hermosilla2021intrinsicextrinsic} at fold/superfamily/family levels, (2) Enzyme Reaction Classification (Reaction) \citep{hermosilla2021intrinsicextrinsic}, (3) Gene Ontology Term Prediction (GO) \citep{Gligorijevi2021StructurebasedPF} of three sub-ontologies: biological process (BP), molecular function (MF), and cellular component (CC), (4) Enzyme Commission (EC) \citep{Gligorijevi2021StructurebasedPF} number prediction, using established dataset splits \citep{fan2023continuousdiscrete}.

\noindent\textbf{Evaluation Metrics.}
Following \citep{fan2023continuousdiscrete}, we use Top-1 accuracy for single-label tasks (FOLD, Reaction); and $F_{max}$ for multi-label tasks (GO, EC), computing the optimal F1-score across prediction thresholds per protein.

\noindent\textbf{Baselines.} 
We select two types of baselines:
\textit{1) protein representation learning methods}:
3DCNN \citep{Derevyanko2018DeepCN},
GraphQA \citep{Baldassarre2020GraphQAPM},
ProtBERT-BFD \citep{Elnaggar2020ProtTransTC},
LM-GVP \citep{Wang2021LMGVPAG}, 
DeepFRI \citep{Gligorijevi2021StructurebasedPF}, 
GVP \citep{jing2021learning},
IEConv \citep{Hermosilla2022ContrastiveRL},
ProNet \citep{wang2023learning},
GearNet \citep{zhang2023protein},
ESM-2 \citep{doi:10.1126/science.ade2574},
CDConv \citep{fan2023continuousdiscrete}, and
EPGGCL \citep{wang2025enhancing};
\textit{2) protein graph construction strategies}:
KNN \citep{jamasb2024evaluating}, Radius \citep{guo2025boosting}, Delaunay triangulation \citep{Khade2023MixedSA}, and Chemical bond \citep{NEURIPS2022_ade039c1}.
For fairness, the graph encoder architecture in MMPG is kept consistent with baselines, differing only in the graph construction strategy.

\noindent\textbf{Details of Training Setup.} 
MMPG is trained on RTX 3090 GPUs.
We use SGD (momentum=0.9, weight decay=5e-4) with an initial LR=1e-2 and a multi-step scheduler.
All results are averaged over 5 random seeds. See code for details.

\subsection{Quantitative Results}
Performance comparison across four protein tasks is shown in Table~\ref{tab:comparison}. 
We have the following observations:
1) MMPG demonstrates superior performance across most tasks, particularly on complex functional prediction. Notably, MMPG achieves a best score of 0.893 on EC and 0.489 on GO-CC. This suggests that MMPG accurately models enzymatic function and cellular roles by integrating diverse information sources (physical, chemical, and geometric).
While ESM-2 excels in FOLD-Superfamily, an evolutionary task, MMPG dominates Fold and Family levels, highlighting its superior capability of precise geometric and structural characterization.
2) MMPG significantly outperforms single-perspective graph construction strategies, demonstrating that combining physical, chemical, and geometric information yields more expressive protein representations.

\subsection{Ablation Study}
Ablation study in Table \ref{tab:ablation} validates each design of MMPG. We have the following observations:
\noindent 1) Each perspective provides unique and task-critical information. Removing any perspective leads to a significant performance drop.
2) MoE proves more effective than other fusion strategies. Edge stack performs poorly (e.g., 0.794 on EC task) as dense, conflicting connections obscure structural patterns, while simple concatenation fails to adequately model inter-perspective interactions, resulting in poor performance on GO tasks.
3) The auxiliary supervision $\mathcal{L}_{\text{CLS\_aux}}$ provides beneficial refinement.
By direct alignment with downstream tasks, $\mathcal{L}_{\text{CLS\_aux}}$ ensures that the gating network receives task-informative representations, thereby reducing noise from irrelevant features to facilitate reliable expert routing.

\subsection{Parameter Analysis}
We analyze the sensitivity of MMPG to key hyperparameters, grouped into multi-perspective protein graph construction and MoE, with results shown in Figure~\ref{fig:param}.

\noindent\textbf{Graph Construction Hyperparameters.} 
Results (Figure~\ref{fig:param} (a), (b), and (c)) indicate that the optimal graph construction hyperparameters align with the level of informational granularity required by each task.
For the energy threshold $\tau$, optimal values range from $\text{-2.0 (for EC)}$ to $\text{-0.5}$ (for GO).
EC hinges on identifying the enzyme's active site.
A strict energy threshold filters out weak interactions, retaining structurally stable regions (e.g., enzyme active sites) crucial for catalytic function~\citep{ribeiro2018mechanism}.
In contrast, GO often describes systemic functions involving multi-domain collaboration.
Looser energy threshold preserves the weaker, long-range contacts that form the interaction pathways essential for these synergistic effects~\citep{sanyal2012long}.
For the similarity, the optimal neighbor count $k$ is 20 for EC and FOLD, and increases to 30 for GO and Reaction. 
EC and FOLD focus on specific residue combinations (e.g., the catalytic triad in EC), requiring strict similarity filtering~\citep{martin1998protein}, whereas GO and Reaction involve broad functional categories or types of reactions, benefiting from capturing more diverse residues~\citep{ashburner2000gene}.
For the radius threshold $r$, a small radius ($r<4$) cannot capture sufficient structural information for robust geometric encoding, while a large radius ($r>4$) decreases the discriminativeness of local spatial patterns by including structurally irrelevant residues, thus leading to inferior performance across tasks.

\noindent\textbf{MoE Hyperparameters.} 
Results are shown in Figure~\ref{fig:param} (d), (e), and (f). 
We first explore the optimal total number of experts $M$, by keeping the ratio of selected experts $K$ fixed at 40\%. 
Performance peaked at $M$ = 10 for most tasks (GO, FOLD, and EC), suggesting this provides a balance between expert diversity and model complexity.
The performance degradation beyond $M$ = 10 likely stems from expert underutilization: each expert receives proportionally fewer samples, resulting in insufficient specialization.
With the expert pool fixed at $M$ = 10, we analyzed the optimal number of selected experts $K$.
We find an optimal range where $K$ is 4 or 6.
A small $K$ (e.g., 2) is likely insufficient to draw knowledge from these perspectives for a given input; while a large $K$ (e.g., 8) may force the gating network to select less relevant experts, harming the precision of the fusion.
This reflects a trade-off between coverage and specialization. 
Finally, we analyze the weight of the auxiliary loss $\lambda$. 
Most tasks achieve optimal performance with a small weight of $\lambda$ = 0.1, but FOLD task benefits from a larger weight of $\lambda$ = 0.5.
Strong auxiliary supervision provides a direct learning signal, suggesting that FOLD's expert routing requires explicit task guidance rather than implicit learning.

\subsection{Analysis of Expert Specialization}

\begin{figure}[t]
\includegraphics[width=0.47\textwidth]{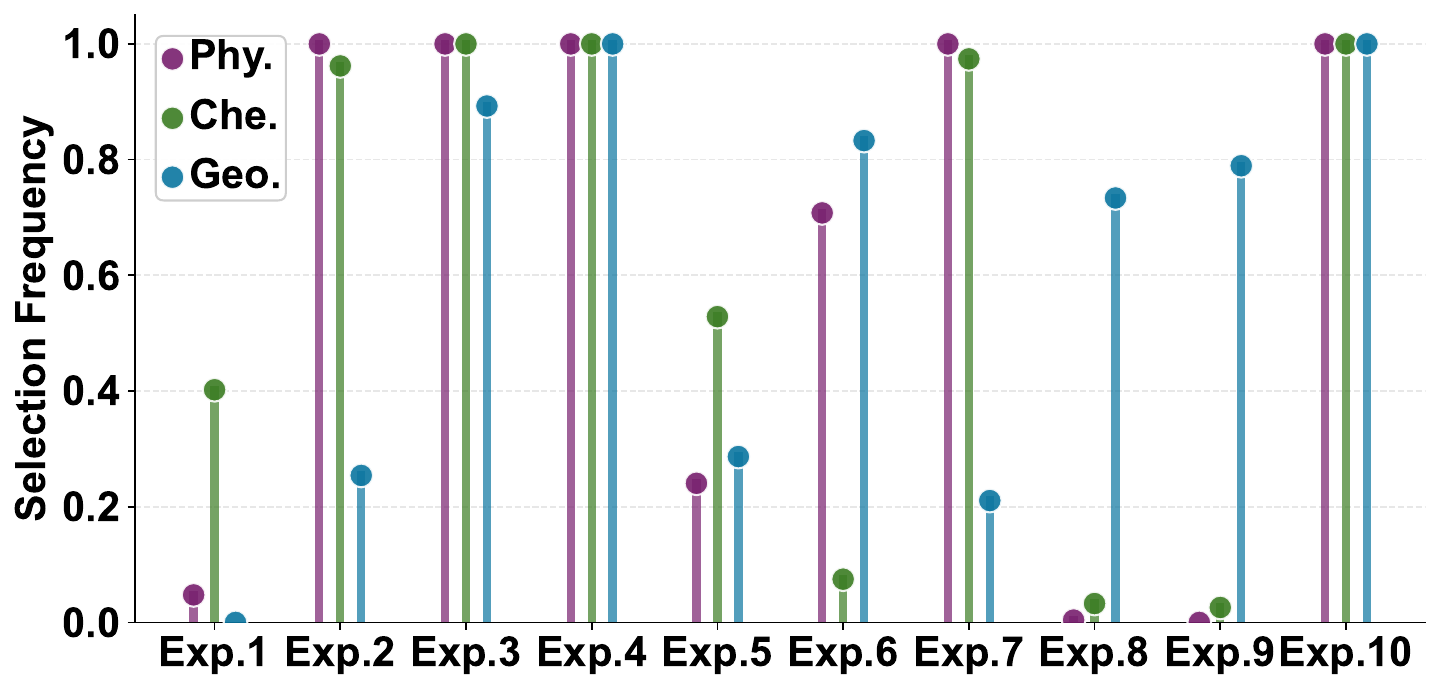}
\centering
\caption{
Expert selection frequency of MoE module for each input perspective on the FOLD task.
}
\label{fig:expert_selection}
\end{figure}

The expert selection patterns of the MoE module provide insights into interactions among different semantic perspectives.
For a detailed analysis, we investigate the FOLD task under the optimal hyperparameter configuration, where $K=6$ experts are selected from a pool of $M=10$.
Figure~\ref{fig:expert_selection} shows that the MoE learns a sophisticated, structured division of labor, manifested as three distinct expert-role types:
1) generalist experts (Experts 3, 4, and 10) are consistently utilized across all perspectives;
2) collaborative experts (Experts 2, 6, and 7) are utilized for specific perspective pairs;
3) specialized experts (Experts 1, 8, and 9) are used for a single perspective.
These patterns reveal a hierarchical organization of interactions among perspectives with three distinct levels:
At the global level, all perspectives are correlated through fundamental protein properties, such as the intrinsic relationship between structure and function.
At the intermediate level, certain perspective pairs exhibit intrinsic interactions.
For example, the physical and geometric perspectives both characterize protein structural properties.
At the granular level, each perspective preserves its own irreplaceable semantic information that cannot be substituted or inferred from others.
Overall, MoE can uncover and disentangle deep inter‑perspective interactions, yielding a more informative protein representation.

\begin{figure}[h]
 \includegraphics[width=0.47\textwidth]{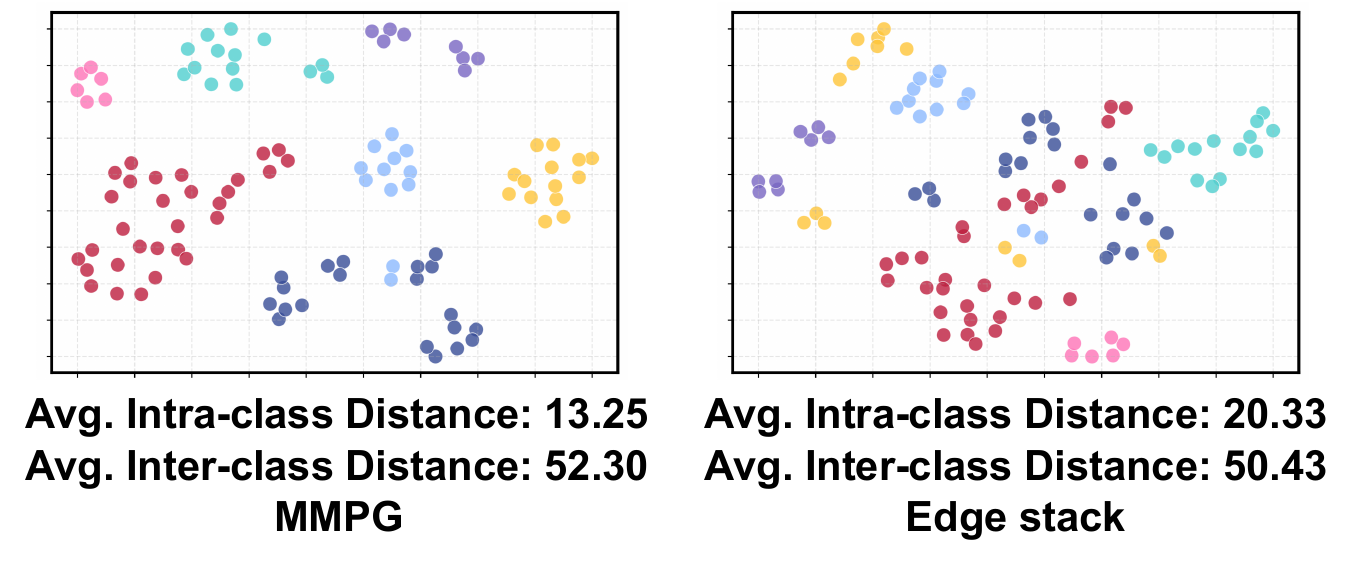}
\centering
\caption{
UMAP projection of learned protein representations with quantified intra-class and inter-class distances. 
}
\label{fig:umap}
\end{figure}

\subsection{Visualization}
We use UMAP projection \citep{mcinnes2018umap} to visualize learned representations for the Reaction task, comparing MMPG with edge stack fusion strategy to assess representation quality.
To demonstrate the results clearly, we randomly select seven classes from the Reaction task. 
As visualized in Figure~\ref{fig:umap}, MMPG exhibits distinct, compact clusters with clear boundaries, while edge-stacking yields scattered distributions with significant class overlap. 
This superior clustering (quantified by tighter intra-class distance: 13.25 vs 20.33 and better inter-class separation: 52.30 vs 50.43) demonstrates MoE's ability to selectively combine perspectives through expert routing, whereas the edge stack strategy indiscriminately superimposes all perspective-specific edges, introducing noise from redundant or conflicting connections that degrade representation quality.

\begin{figure}[h]
 \includegraphics[width=0.46\textwidth]{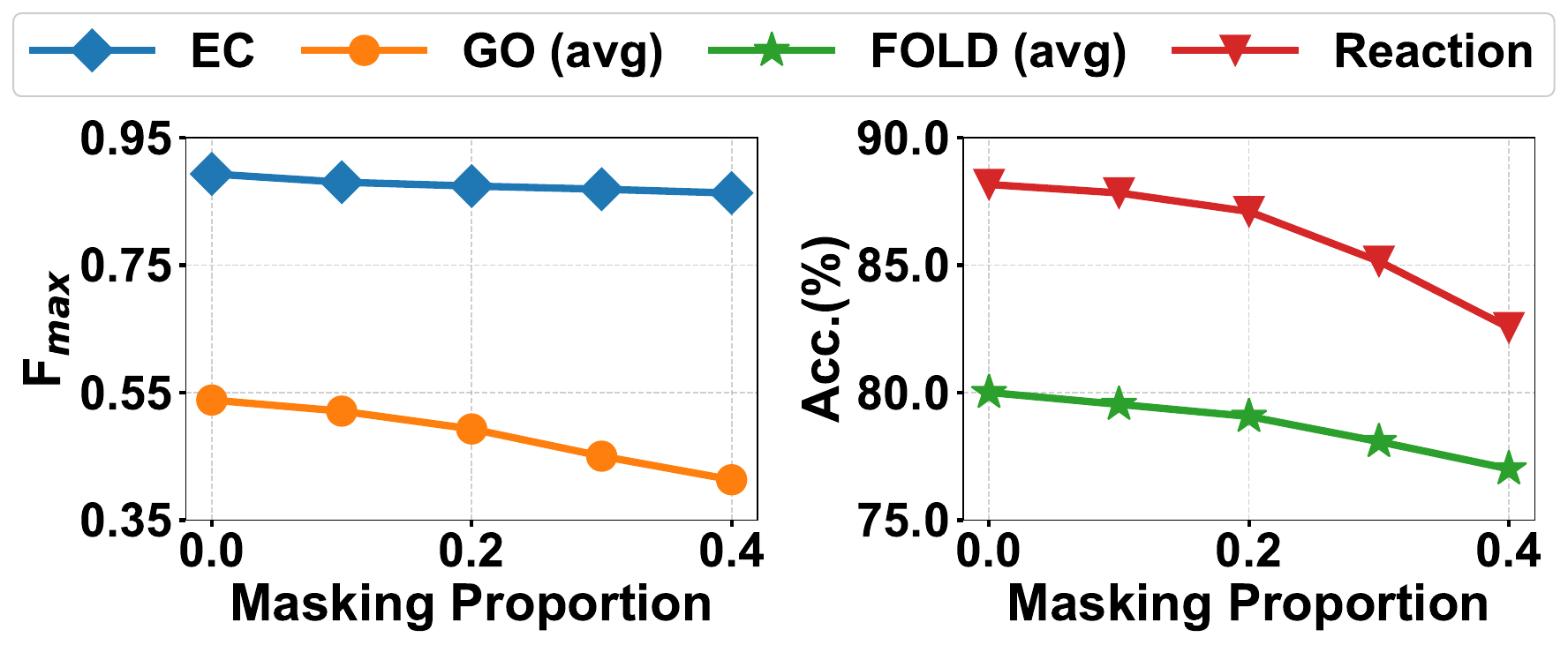}
\centering
\caption{
Robustness analysis of MMPG under random residue masking (0-40\%) across four protein tasks.
}
\label{fig:robustness}
\end{figure}

\subsection{Robustness Analysis}
We evaluate the robustness of MMPG under conditions mimicking low-resolution experimental data by randomly masking u\% of input residue embeddings, reflecting scenarios where residue chemical identities are uncertain or unresolved \citep{dimaio2009refinement}.
As shown in Figure~\ref{fig:robustness}, MMPG demonstrates strong resilience. 
Even with 40\% masking, performance degradation remains modest, particularly for EC and FOLD tasks. 
This robustness might stem from the multi-perspective design.
MMPG can reconstruct missing information by leveraging the diverse information patterns that remain visible across other perspectives.

\section{Conclusion}
This work addresses the limitations of single-perspective protein graph construction in capturing comprehensive protein information. 
We build physical, chemical, and geometric protein graphs for a comprehensive representational foundation.
Building upon this, we design an MoE module that discovers and leverages deep interactions among perspectives, dynamically integrating them into task-specific representations. Our method achieves advanced performance across different protein downstream tasks, demonstrating its effectiveness in PRL.

\section{Acknowledgments}
This work was supported by Government Special Support Funds for the Guangdong Institute of Intelligence Science and Technology, and National Natural Science Foundation of China (NSFC) under Grant No.62506084.

\bigskip

\bibliography{aaai2026}

\end{document}